\documentclass{article}

\usepackage{microtype}
\usepackage{graphicx}
\usepackage{subfigure}
\usepackage{booktabs} 
\usepackage{amsfonts}
\usepackage{amsmath}
\usepackage{makecell}
\usepackage{siunitx}
\usepackage{hyperref}

\usepackage[accepted]{icml2021}

\icmltitlerunning{Autoregressive Denoising Diffusion Models for Multivariate Probabilistic Time Series Forecasting}

\begin{document}

\twocolumn[
\icmltitle{Autoregressive Denoising Diffusion Models for Multivariate Probabilistic Time Series Forecasting}




\begin{icmlauthorlist}
\icmlauthor{Kashif Rasul}{zr}
\icmlauthor{Calvin Seward}{zr}
\icmlauthor{Ingmar Schuster}{zr}
\icmlauthor{Roland Vollgraf}{zr}
\end{icmlauthorlist}

\icmlaffiliation{zr}{Zalando Research,  Mühlenstraße 25, 10243 Berlin, Germany}

\icmlcorrespondingauthor{Kashif Rasul}{kashif.rasul@zalando.de}

\icmlkeywords{Time Series and Sequences, Generative Models}

\vskip 0.3in
]



\printAffiliationsAndNotice{} 

\begin{abstract}
In this work, we propose \texttt{TimeGrad}, an autoregressive model for multivariate probabilistic time series forecasting which samples from the data distribution at each time step by estimating its gradient. To this end, we use diffusion probabilistic models, a class of latent variable models closely connected to score matching and energy-based methods. Our model learns gradients by optimizing a variational bound on the data likelihood and at inference time converts white noise into a sample of the distribution of interest through a Markov chain using Langevin sampling. We demonstrate experimentally that the proposed autoregressive denoising diffusion model is the new state-of-the-art multivariate probabilistic forecasting method on real-world data sets with thousands of correlated dimensions. We hope that this method is a useful tool for practitioners and lays the foundation for future research in this area.
\end{abstract}

\section{Introduction}
\label{sec:intro}

Classical time series forecasting methods such as those in~\cite{hyndman2018forecasting} typically provide univariate point forecasts, require hand-tuned features to model seasonality, and are trained individually on each time series. Deep learning based time series models \citep{benidis2020neural} are  popular alternatives due to their end-to-end training of a global model, ease of incorporating exogenous covariates, and automatic feature extraction abilities. The task of modeling uncertainties is of vital importance for downstream problems that use these forecasts for (business) decision making. More often the individual time series for a problem data set are statistically dependent on each other. Ideally, deep learning models need to incorporate this inductive bias in the form of  multivariate \citep{tsay} probabilistic methods to provide accurate forecasts. 

To model the full predictive distribution, methods typically resort to tractable distribution classes or some type of low-rank approximations, regardless of the true data distribution. To model the distribution in a general fashion, one needs probabilistic methods with tractable likelihoods. Till now several deep learning methods have been proposed  for this purpose such as autoregressive \cite{pmlr-v48-oord16} or generative ones based on normalizing flows \cite{papamakarios2019normalizing} which can learn flexible models of high dimensional multivariate time series. Even if the full likelihood is not be tractable, one can often optimize a tractable lower bound to the likelihood. But still, these methods require a certain structure in the functional approximators, for example on the determinant of the Jacobian \cite{45819} for normalizing flows. \emph{Energy-based models} (EBM) \cite{10.1162/089976602760128018, lecun-06} on the other hand have a much less restrictive functional form. They approximate the unnormalized log-probability so that density estimation reduces to a non-linear regression problem. EBMs have been shown to perform well in learning high dimensional data distributions at the cost of being difficult to train \cite{Song2021HowTT}.

In this work, we propose autoregressive EBMs to solve the multivariate probabilistic time series forecasting problem via a model we call \texttt{TimeGrad} and show that not only are we able to train such a model with all the inductive biases of probabilistic time series forecasting, but this model performs exceptionally well when compared to other modern methods.  This autoregressive-EBM combination retains the power of autoregressive models, such as good performance in extrapolation into the future, with the flexibility of EBMs as a general purpose high-dimensional distribution model, while remaining computationally tractable.

The paper is organized as follows. In Section~\ref{sec:Background} we first set up the notation and detail the EBM of  \cite{ho2020denoising} which forms the basis of our per time-step distribution model.  Section~\ref{sec:Model} introduces the multivariate probabilistic time series problem and we detail the \texttt{TimeGrad} model. The experiments with extensive results are detailed in Section~\ref{sec:Experiments}. We  cover related work in Section~\ref{sec:Related} and conclude with some discussion in Section~\ref{sec:Conclusion}. 

\section{Diffusion Probabilistic Model}
\label{sec:Background}

Let $\mathbf{x}^0 \sim q_{\mathcal{X}}(\mathbf{x}^0)$ denote the multivariate training vector from some input space $\mathcal{X} = \mathbb{R}^D$ and let $p_\theta(\mathbf{x}^0)$ denote the probability density function (PDF) which aims to approximate $q_{\mathcal{X}}(\mathbf{x}^0)$ and allows for easy sampling.  Diffusion models \cite{pmlr-v37-sohl-dickstein15} are latent variable models of the form $p_\theta(\mathbf{x}^0) := \int p_\theta(\mathbf{x}^{0:N}) \, \mathrm{d}\mathbf{x}^{1:N}$, where $\mathbf{x}^{1}, \ldots, \mathbf{x}^N$ are latents of dimension $\mathbb{R}^D$. 
 Unlike in variational autoencoders  \cite{DBLP:journals/ftml/KingmaW19} the approximate posterior $q(\mathbf{x}^{1:N} | \mathbf{x}^{0})$, 
 $$
 q(\mathbf{x}^{1:N} | \mathbf{x}^{0}) = \Pi_{n=1}^{N}  q(\mathbf{x}^{n} | \mathbf{x}^{n-1}) 
 $$
 is not trainable but fixed to a Markov chain (called the \emph{forward} process) that gradually adds Gaussian noise to the signal:
 $$
 q(\mathbf{x}^{n} | \mathbf{x}^{n-1}) := \mathcal{N}(\mathbf{x}^{n}; \sqrt{1 - \beta_n} \mathbf{x}^{n-1}, \beta_n \mathbf{I}).
 $$
 The forward process uses an increasing variance schedule $\beta_1, \ldots, \beta_{N}$ with $\beta_n \in (0,1)$.
 The joint distribution $ p_\theta(\mathbf{x}^{0:N})$  is called the \emph{reverse} process, and is defined as a Markov chain with learned Gaussian transitions starting with $p(\mathbf{x}^N) = \mathcal{N}(\mathbf{x}^N; \mathbf{0}, \mathbf{I})$, where each subsequent transition of
\begin{equation*}
     p_\theta(\mathbf{x}^{0:N}) := p(\mathbf{x}^N) \Pi_{n=N}^1 p_\theta(\mathbf{x}^{n-1} | \mathbf{x}^{n})
\end{equation*}
is given by a parametrization of our choosing denoted by
 \begin{equation}
 \label{eqn:parametrization}
     p_\theta(\mathbf{x}^{n-1} | \mathbf{x}^{n}) :=  \mathcal{N}(\mathbf{x}^{n-1}; \mu_{\theta}(\mathbf{x}^{n},n), \Sigma_\theta(\mathbf{x}^{n},n) \mathbf{I}),
 \end{equation}
with shared parameters $\theta$. Both $\mu_\theta: \mathbb{R}^D \times \mathbb{N} \to \mathbb{R}^D $ and $\Sigma_\theta : \mathbb{R}^D \times \mathbb{N} \to \mathbb{R}^{+}$ take two inputs, namely the variable $\mathbf{x}^n \in \mathbb{R}^D$ as well as the noise index $n \in \mathbb{N}$.
The goal of $ p_\theta(\mathbf{x}^{n-1} | \mathbf{x}^{n})$ is to eliminate the Gaussian noise added in the diffusion process.  The parameters $\theta$ are learned to fit the data distribution $q_{\mathcal{X}}(\mathbf{x}^0)$ by minimizing the negative log-likelihood via a variational  bound using Jensen's inequality:
\begin{equation*}
\begin{split}
\min_{\theta} \mathbb{E}_{q(\mathbf{x}^0)}[- \log p_\theta(\mathbf{x}^0)]   \leq \\ \min_{\theta}  \mathbb{E}_{q(\mathbf{x}^{0:N})}[ - \log p_\theta(\mathbf{x}^{0:N}) + \log q(\mathbf{x}^{1:N} | \mathbf{x}^0)].
\end{split}
\end{equation*}
This upper bound can be shown to be equal to
\begin{equation}
    \label{eqn:label}
    \min_{\theta}  \mathbb{E}_{q(\mathbf{x}^{0:N})} \left[ - \log p(\mathbf{x}^N)   - \sum_{n=1}^{N} \log  \frac{p_\theta(\mathbf{x}^{n-1} | \mathbf{x}^{n})}{ q(\mathbf{x}^{n} | \mathbf{x}^{n-1})} \right].
\end{equation}
As shown by  \cite{ho2020denoising}, a property of the forward process is that it admits sampling $\mathbf{x}^n$ at any arbitrary noise level $n$ in closed form, since if $\alpha_n := 1 - \beta_n$ and $\bar{\alpha}_n := \Pi_{i=1}^n \alpha_i$ its cumulative product, we have:
\begin{equation}
\label{eqn:prop}
q(\mathbf{x}^{n} | \mathbf{x}^{0}) =  \mathcal{N}(\mathbf{x}^{n}; \sqrt{\bar{\alpha}_n} \mathbf{x}^{0}, (1 - \bar{\alpha}_n) \mathbf{I}).
\end{equation} 
By using the fact that these processes are Markov chains, the objective in (\ref{eqn:label}) can be written as the KL-divergence between Gaussian distributions:
\begin{multline}
\label{eqn:objective-full}
    - \log p_\theta(\mathbf{x}^0| \mathbf{x}^1) + D_{\mathrm{KL}}( q(\mathbf{x}^N | \mathbf{x}^0) || p(\mathbf{x}^N)) \\
    + \sum_{n=2}^{N} D_{\mathrm{KL}}( q(\mathbf{x}^{n-1}| \mathbf{x}^n, \mathbf{x}^0) || p_\theta(\mathbf{x}^{n-1} | \mathbf{x}^n)),
\end{multline}
and \cite{{ho2020denoising}} shows that by the property (\ref{eqn:prop}) the forward process posterior in these KL divergences when conditioned on $\mathbf{x}^0$, i.e. $q(\mathbf{x}^{n-1}| \mathbf{x}^n, \mathbf{x}^0)$ are tractable given by
$$
q(\mathbf{x}^{n-1}| \mathbf{x}^n, \mathbf{x}^0) = {\cal{N}}(\mathbf{x}^{n-1}; \tilde{\mu}_{n}(\mathbf{x}^n, \mathbf{x}^0), \tilde{\beta}_n\mathbf{I}),
$$
where
$$ 
\tilde{\mu}_{n}(\mathbf{x}^n, \mathbf{x}^0) := \frac{\sqrt{\bar{\alpha}_{n-1} }\beta_n}{1 - \bar{\alpha}_n} \mathbf{x}^0 + \frac{\sqrt{\alpha_n} (1- \bar{\alpha}_{n-1})} {1 - \bar{\alpha}_n} \mathbf{x}^n
$$
and
\begin{equation}
\label{eqn:beta-tilde}
\tilde{\beta}_n := \frac{1 - \bar{\alpha}_{n-1}}{1 - \bar{\alpha}_n} \beta_n.
\end{equation}
Further, \cite{{ho2020denoising}} shows that the KL-divergence between Gaussians can be written as:
\begin{multline}
D_{\mathrm{KL}}( q(\mathbf{x}^{n-1}| \mathbf{x}^n, \mathbf{x}^0) || p_\theta(\mathbf{x}^{n-1} | \mathbf{x}^n)) = \\
\mathbb{E}_{q} \left[ \frac{1}{2 \Sigma_\theta} \| \tilde{\mu}_n(\mathbf{x}^n, \mathbf{x}^0) - \mu_\theta(\mathbf{x}^n, n)\|^2 \right] + C, 
\end{multline}
where $C$ is a constant which does not depend on $\theta$. So instead of a parametrization (\ref{eqn:parametrization}) of $p_\theta$ that predicts $\tilde{\mu}$, one can instead use the property (\ref{eqn:prop}) to write $\mathbf{x}^n(\mathbf{x}^0, \mathbf{\epsilon}) = \sqrt{\bar{\alpha}_n} \mathbf{x}^0 + \sqrt{1 - \bar{\alpha}_n} \mathbf{\epsilon}$ for $\mathbf{\epsilon} \sim {\cal{N}}(\mathbf{0}, \mathbf{I})$ and the formula for $\tilde{\mu}$ to obtain that $\mu_\theta$ must predict $(\mathbf{x}^n - \beta_n \mathbf{\epsilon} / \sqrt{1 - \bar{\alpha}_n})/\sqrt{\alpha_n}$, but since $\mathbf{x}^n$ is available to the network, we can choose:
\begin{equation*}
    \mu_\theta(\mathbf{x}^n, n) = \frac{1}{\sqrt{\alpha_n}} \left( \mathbf{x}^n - \frac{\beta_n}{\sqrt{1 - \bar{\alpha}_n}} \mathbf{\epsilon}_\theta( \mathbf{x}^n, n)\right),
\end{equation*}
where $\mathbf{\epsilon}_\theta$ is a network which predicts $\mathbf{\epsilon} \sim {\cal{N}}(\mathbf{0}, \mathbf{I})$ from $\mathbf{x}^n$, so that the objective simplifies to:
\begin{equation}
\label{eqn:objective}
    \mathbb{E}_{\mathbf{x}^0, \mathbf{\epsilon}} \left[ \frac{\beta_n^2}{2 \Sigma_\theta \alpha_n (1 - \bar{\alpha}_n)}  \| \mathbf{\epsilon} - \mathbf{\epsilon}_\theta (\sqrt{\bar{\alpha}_n} \mathbf{x}^0 + \sqrt{1 - \bar{\alpha}_n} \mathbf{\epsilon}, n ) \|^ 2\right]
\end{equation}
resembling the loss in Noise Conditional Score Networks \cite{NEURIPS2019_3001ef25, NEURIPS2020_song_erman} using score matching. Once trained, to sample from the reverse process $\mathbf{x}^{n-1} \sim p_\theta(\mathbf{x}^{n-1} | \mathbf{x}^n)$  (\ref{eqn:parametrization}) we can compute 
$$
\mathbf{x}^{n-1} = \frac{1}{\sqrt{\alpha_n}} \left(\mathbf{x}^n - \frac{\beta_n}{\sqrt{1 - \bar{\alpha}_n}} \mathbf{\epsilon}_\theta(\mathbf{x}^n, n) \right) + \sqrt{\Sigma_\theta} \mathbf{z}
$$
where $\mathbf{z} \sim {\cal{N}}(\mathbf{0}, \mathbf{I})$ for $n = N, \ldots, 2$ and $\mathbf{z} = \mathbf{0}$ when $n=1$. The full sampling procedure for $\mathbf{x}^0$, starting from white noise sample $\mathbf{x}^N$, resembles Langevin dynamics where we sample from the most noise-perturbed distribution and reduce the magnitude of the noise scale until we reach the smallest one.

\section{\texttt{TimeGrad} Method}
\label{sec:Model}

We denote the entities of a multivariate time series by  $x_{i,t}^0 \in \mathbb{R}$  for $i \in \{1,\ldots,D\}$  where $t$ is the time index. Thus the multivariate vector at time $t$ is given by $\mathbf{x}_t^0 \in \mathbb{R}^D$. We are tasked with predicting the multivariate distribution some given prediction time steps into the future and so in what follows consider time series with $t \in [1, T]$,  sampled from the complete time series history of the  training data, where we will split this contiguous sequence into a context window of size $[1,t_0)$ and prediction interval $[t_0, T]$, reminiscent of seq-to-seq models \cite{NIPS2014_5346} in language modeling.

In the univariate probabilistic \texttt{DeepAR} model \citep{DBLP:journals/corr/FlunkertSG17}, the log-likelihood of each entity $x^0_{i,t}$ at a time step $t \in [t_0, T]$ is maximized over an individual time series' prediction window. This is done with respect to the  parameters of some chosen distributional model via the state of an RNN derived from its previous time step $x^0_{i,t-1}$ and  its corresponding covariates $\mathbf{c}_{i,t-1}$. The emission distribution model, which is typically Gaussian for real-valued data or negative binomial for count data, is selected to best match the statistics of the time series and the network incorporates activation functions that satisfy the constraints of the distribution's parameters, e.g. a \texttt{softplus()} for the scale parameter of the Gaussian.

A straightforward time series model for multivariate real-valued data could use a factorizing output distribution instead. Shared parameters can then learn patterns across the individual time series entities through the temporal component --- but the model falls short of capturing dependencies in the emissions of the model. For this, a full joint distribution at each time step has to be modeled, for example by using a  multivariate Gaussian. However, modeling the full covariance matrix not only increases the number of parameters of the neural network by $O(D^2)$, making learning difficult but computing the loss is $O(D^3)$ making it impractical. Furthermore, statistical dependencies for such  distributions would be limited to second-order effects. Approximating Gaussians with low-rank covariance matrices do work however and these models are referred to as \texttt{Vec-LSTM} in~\cite{NIPS2019_8907}.

\begin{figure}[ht]
\vskip 0.2in
\begin{center}
\centerline{\includegraphics[width=\columnwidth]{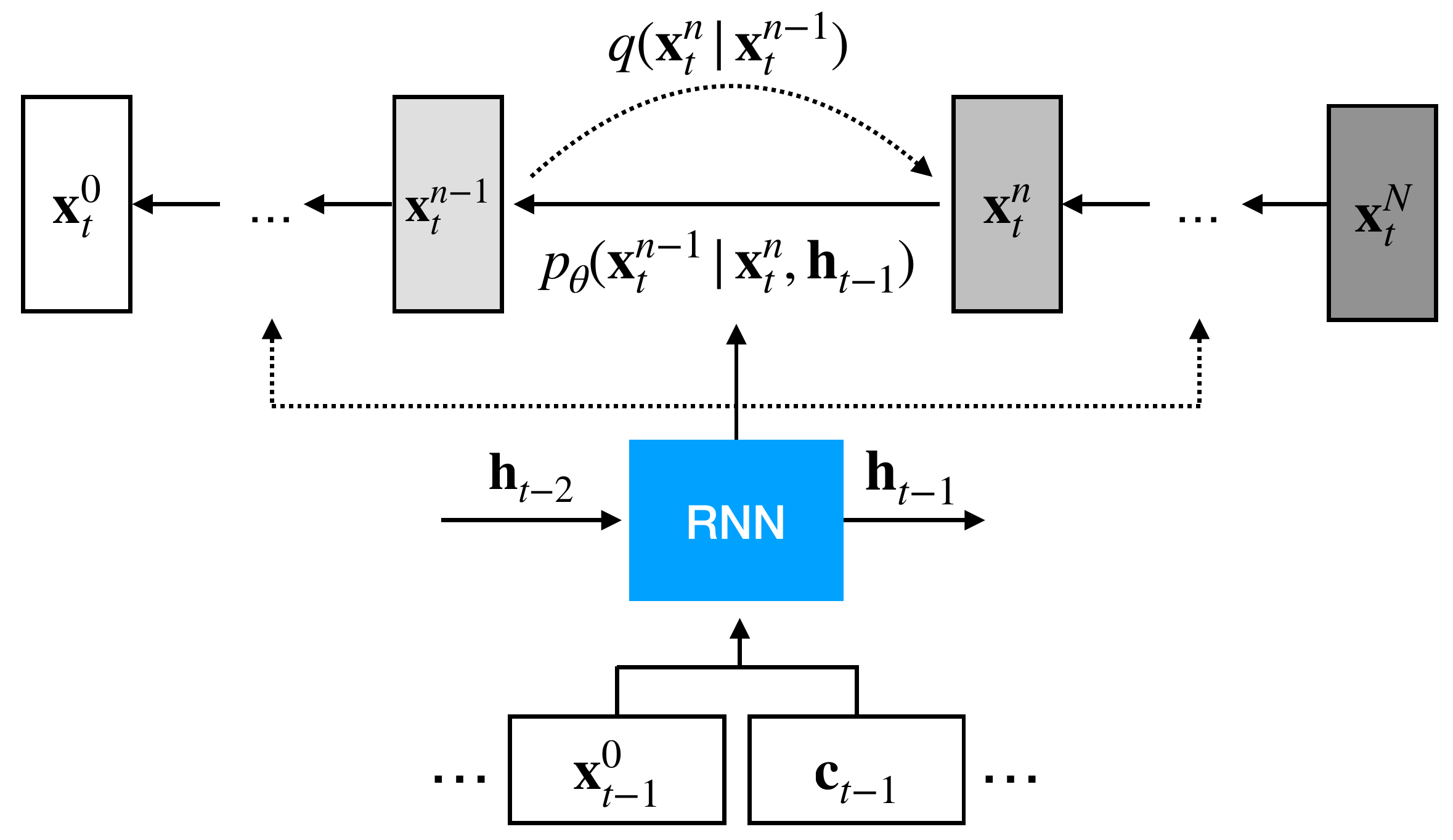}}
\caption{\texttt{TimeGrad} schematic: an RNN \emph{conditioned} diffusion probabilistic model  at some time $t-1$ depicting the fixed forward process that adds Gaussian noise  and the learned  reverse processes.}
\label{time-grad-fig}
\end{center}
\vskip -0.2in
\end{figure}

Instead, in this work we propose \texttt{TimeGrad} which aims to learn a  model of the conditional distribution of the future time steps of a multivariate time series given its past and covariates as:
\begin{equation}
\label{model}
q_{\mathcal{X}}(\mathbf{x}_{t_0:T}^0 | \mathbf{x}_{1:t_0-1}^0, \mathbf{c}_{1:T}) = \Pi_{t=t_0}^{T} q_{\mathcal{X}}(\mathbf{x}_{t}^0 | \mathbf{x}_{1:t-1}^0, \mathbf{c}_{1:T}),
\end{equation}
were we assume that the covariates are known for all the time points and each factor is learned via a \emph{conditional} denoising diffusion model introduced above. To model the temporal dynamics we employ the autoregressive recurrent neural network (RNN) architecture from \cite{Graves13, NIPS2014_5346} which utilizes the LSTM \cite{6795963} or GRU \cite{69e088c8129341ac89810907fe6b1bfe} to encode the time series sequence up to time point $t$, given the covariates $\mathbf{c}_t$, via the updated hidden state $\mathbf{h}_t$:
\begin{equation}
\label{eqn:rnn}
\mathbf{h}_t = \mathrm{RNN}_\theta(\mathtt{concat}(\mathbf{x}_{t}^0, \mathbf{c}_{t}), \mathbf{h}_{t-1}),
\end{equation}
where $\mathrm{RNN}_\theta$ is a multi-layer LSTM or GRU parameterized by shared weights $\theta$ and $\mathbf{h}_0 = \mathbf{0}$. Thus we can approximate (\ref{model}) by the model 
\begin{equation}
    \label{cond_model}
    \Pi_{t=t_0}^T p_\theta(\mathbf{x}_{t}^0 | \mathbf{h}_{t-1}),
\end{equation}
where now $\theta$ comprises the weights of the RNN as well as denoising diffusion model. This model is autoregressive as it consumes the observations at the time step $t-1$ as input to learn the distribution of, or sample,  the next time step  as shown in Figure \ref{time-grad-fig}.

\subsection{Training}

Training is performed by randomly sampling context and adjoining prediction sized windows from the training time series data and optimizing the parameters $\theta$ that minimize the negative log-likelihood of the model (\ref{cond_model}):
$$
 \sum_{t=t_0}^T - \log p_\theta(\mathbf{x}_t^0 | \mathbf{h}_{t-1}),
$$
starting with the hidden state $\mathbf{h}_{t_0-1}$ obtained by running the RNN on the chosen context window. Via a similar derivation as in the previous section, we have that the  conditional variant of the objective (\ref{eqn:objective-full})  for  time step $t$ and noise index $n$ is  given by the following simplification of (\ref{eqn:objective}) \cite{ho2020denoising}:
$$
 \mathbb{E}_{\mathbf{x}_t^0, \epsilon, n} \left[ \| \mathbf{\epsilon} - \mathbf{\epsilon}_\theta (\sqrt{\bar{\alpha}_n} \mathbf{x}^0_t + \sqrt{1 - \bar{\alpha}_n} \mathbf{\epsilon}, \mathbf{h}_{t-1}, n ) \|^ 2 \right],
$$
when we choose the variance in (\ref{eqn:parametrization}) to be $\Sigma_\theta = \tilde{\beta}_n$ (\ref{eqn:beta-tilde}), where now the $\epsilon_\theta$ network is also \emph{conditioned} on the hidden state. Algorithm \ref{alg:training} is the training procedure for each time step in the prediction window using this objective.

\begin{algorithm}[tb]
  \caption{Training for each time series step $t \in [t_0, T]$}
  \label{alg:training}
\begin{algorithmic}
  \STATE {\bfseries Input:} data $\mathbf{x}_t^0 \sim q_{\cal{X}}(\mathbf{x}_t^0)$ and  state $\mathbf{h}_{t-1}$
  \REPEAT
  \STATE Initialize $n \sim \mathrm{Uniform}({1, \ldots, N})$ and $\epsilon \sim {\cal{N}}(\mathbf{0}, \mathbf{I})$
  Take gradient step on\;
  $$ \nabla_\theta \| \mathbf{\epsilon} - \mathbf{\epsilon}_\theta (\sqrt{\bar{\alpha}_n} \mathbf{x}^0_t + \sqrt{1 - \bar{\alpha}_n} \mathbf{\epsilon}, \mathbf{h}_{t-1}, n ) \|^ 2$$
  \UNTIL{converged}
\end{algorithmic}
\end{algorithm}

\subsection{Inference}

After training, we wish to predict for each time series in our data set some prediction steps into the future and compare with the corresponding test set time series. As in training, we run the RNN over the last context sized window of the training set to obtain the hidden state $\mathbf{h}_{T}$ via (\ref{eqn:rnn}). Then we follow the sampling procedure in Algorithm \ref{alg:sampling} to obtain a sample  $\mathbf{x}_{T+1}^0$ of the next time step, which we can pass autoregressively to the RNN together with the covariates $\mathbf{c}_{T+1}$ to obtain the next hidden state $\mathbf{h}_{T+1}$ and repeat until the  desired forecast horizon has been reached. This process of sampling trajectories from the ``warm-up'' state $\mathbf{h}_{T}$ can be repeated many times (e.g. $S=100$)  to obtain empirical quantiles of the uncertainty of our predictions.

\begin{algorithm}[tb]
  \caption{Sampling  $\mathbf{x}_t^0$ via annealed Langevin dynamics} 
  \label{alg:sampling}
\begin{algorithmic}
  \STATE {\bfseries Input:} noise $\mathbf{x}_t^N \sim {\cal{N}}(\mathbf{0}, \mathbf{I})$ and state  $\mathbf{h}_{t-1}$ 
    \FOR{$n=N$ {\bfseries to} $1$}
        \IF{$n > 1$} 
         \STATE $\mathbf{z} \sim {\cal{N}}(\mathbf{0}, \mathbf{I})$
         \ELSE
         \STATE $\mathbf{z} = \mathbf{0}$
        \ENDIF
        \STATE $\mathbf{x}_t^{n-1} = \frac{1}{\sqrt{\alpha_n}} (\mathbf{x}^n_t - \frac{\beta_n}{\sqrt{1 - \bar{\alpha}_n}} \mathbf{\epsilon}_\theta(\mathbf{x}^n_t, \mathbf{h}_{t-1}, n)) + \sqrt{\Sigma_\theta} \mathbf{z}$
    \ENDFOR
    \STATE \textbf{Return:}  $\mathbf{x}_t^0$ 
\end{algorithmic}
\end{algorithm}

\subsection{Scaling}
In real-world data, the magnitudes of different time series entities can vary drastically. To normalize scales, we divide each time series entity by their context window mean (or $1$ if it's zero) before feeding it into the model. At inference, the samples are then multiplied by the same mean values to match the original scale. This rescaling technique simplifies the problem for the model, which is reflected in significantly improved empirical performance as shown in~\cite{DBLP:journals/corr/FlunkertSG17}. The other method of a short-cut connection from the input to the output of the function approximator, as done in the multivariate point forecasting method \texttt{LSTNet} \cite{Lai:2018:MLS:3209978.3210006}, is not applicable here.

\subsection{Covariates}

We employ embeddings for categorical features~\citep{twimlJD}, that allows for relationships within a category, or its context, to be captured when training time series models. 
Combining these embeddings as features for forecasting yields powerful models like the first place winner of the Kaggle Taxi Trajectory Prediction\footnote{\url{https://www.kaggle.com/c/pkdd-15-predict-taxi-service-trajectory-i}} challenge~\citep{DBLP:journals/corr/BrebissonSAVB15}. The covariates $\mathbf{c}_t$ we use are composed of time-dependent (e.g. day of week, hour of day) and time-independent embeddings, if applicable, as well as lag features depending on the time frequency of the data set we are training on. All covariates are thus known for the  periods we wish to forecast.

\section{Experiments}
\label{sec:Experiments}

We benchmark \texttt{TimeGrad} on \emph{six} real-world data sets and evaluate against several competitive baselines. The source code of the model will be made available after the review process.

\subsection{Evaluation Metric and Data Set}

For evaluation, we compute the  Continuous Ranked Probability Score (CRPS)~\citep{RePEc} on each  time series dimension, as well as on the sum of all time series dimensions (the latter denoted by $\mathrm{CRPS}_{\mathrm{sum}}$). CRPS  measures the compatibility of a cumulative distribution function $F$ with an observation $x$ as
$$
    \mathrm{CRPS}(F, x) = \int_\mathbb{R} (F(z) -  \mathbb{I} \{x \leq z\})^2\, \mathrm{d}z,
$$
where $\mathbb{I}\{ x \leq z\}$ is the indicator function which is one if $x \leq z$ and zero otherwise. CRPS is a \emph{proper scoring function}, hence CRPS attains its minimum when the predictive distribution $F$ and the data distribution are equal.
Employing the empirical CDF of $F$, i.e. $\hat F(z) = \frac{1}{S} \sum_{s=1}^S \mathbb{I}\{X_s \leq z\}$ with $S$ samples $X_s \sim F$ as a natural approximation of the predictive CDF, CRPS can be directly computed from simulated samples of the conditional distribution (\ref{model}) at each time point~\citep{JSSv090i12}. 
Finally, $\mathrm{CRPS}_{\mathrm{sum}}$ is obtained by first summing across the $D$ time-series --- both for the ground-truth data, and sampled data (yielding $\hat F_{\mathrm{sum}}(t)$ for each time point).  
The results are then averaged over the prediction horizon, i.e. formally $\mathrm{CRPS}_{\mathrm{sum}} = \mathbb{E}_t \left[ \mathrm{CRPS} \left( \hat F_{\mathrm{sum}}(t), \sum_i x_{i,t}^0 \right) \right]$. As proved in \cite{NIPS2020_Emmanuel} $\mathrm{CRPS}_{\mathrm{sum}}$ is also a proper scoring function and we use it, instead of likelihood based metrics, since not all methods we compare against yield analytical forecast distributions or likelihoods are not meaningfully defined.

For our experiments we use \texttt{Exchange}~\citep{Lai:2018:MLS:3209978.3210006}, \texttt{Solar}~\citep{Lai:2018:MLS:3209978.3210006},  \texttt{Electricity}\footnote{\url{https://archive.ics.uci.edu/ml/datasets/ElectricityLoadDiagrams20112014}}, \texttt{Traffic}\footnote{\url{https://archive.ics.uci.edu/ml/datasets/PEMS-SF}}, \texttt{Taxi}\footnote{\url{https://www1.nyc.gov/site/tlc/about/tlc-trip-record-data.page}} and \texttt{Wikipedia}\footnote{\url{https://github.com/mbohlkeschneider/gluon-ts/tree/mv_release/datasets}} open data sets, preprocessed exactly as in~\cite{NIPS2019_8907}, with their properties listed in Table~\ref{dataset}. As can be noted in the table, we do not need to normalize scales for \texttt{Traffic}.

\begin{table}[t]
\caption{Dimension, domain, frequency, total training time steps and prediction length properties of the  training data sets used in the experiments.}
\label{dataset}
\vskip 0.15in
\begin{center}
\begin{small}
\begin{sc}
\begin{tabular}{lccccc}
\toprule
Data set &   \makecell{Dim.\\ $D$} & Dom. & Freq. & \makecell{Time \\ steps}  & \makecell{Pred.\\ steps}\\ 
\midrule
\texttt{Exchange}  & $8$ & $\mathbb{R}^{+}$ & day & $6,071$ & $30$\\
\texttt{Solar}    & $137$ & $\mathbb{R}^{+}$ & hour & $7,009$ & $24$\\
 \texttt{Elec.} & $370$ &  $\mathbb{R}^{+}$ & hour & $5,833$ & $24$\\
\texttt{Traffic}  & $963$  & $(0,1)$ & hour & $4,001$ & $24$\\
\texttt{Taxi}    & $1,214$ & $\mathbb{N}$ & 30-min & $1,488$ & $24$\\
\texttt{Wiki.}   & $2,000$ & $\mathbb{N}$ & day & $792$ & $30$\\
\bottomrule
\end{tabular}
\end{sc}
\end{small}
\end{center}
\vskip -0.1in
\end{table}

\subsection{Model Architecture}

We train  \texttt{TimeGrad} via SGD using Adam~\cite{kingma:adam} with learning rate of $\num{1e-3}$ on the training split of each data set with $N=100$ diffusion steps using a linear variance schedule starting from  $\beta_1=\num{1e-4}$ till  $\beta_N = 0.1$. We construct batches of size $64$ by taking random windows (with possible overlaps), with the context size set to the number of prediction steps, from the total time steps of each data set (see Table \ref{dataset}).   For testing we use a rolling windows prediction starting from the last context window  history before the start of the prediction and compare it to the ground-truth in the test set by sampling $S=100$ trajectories.

\begin{figure}[ht]
\vskip 0.2in
\begin{center}
\centerline{\includegraphics[width=\columnwidth]{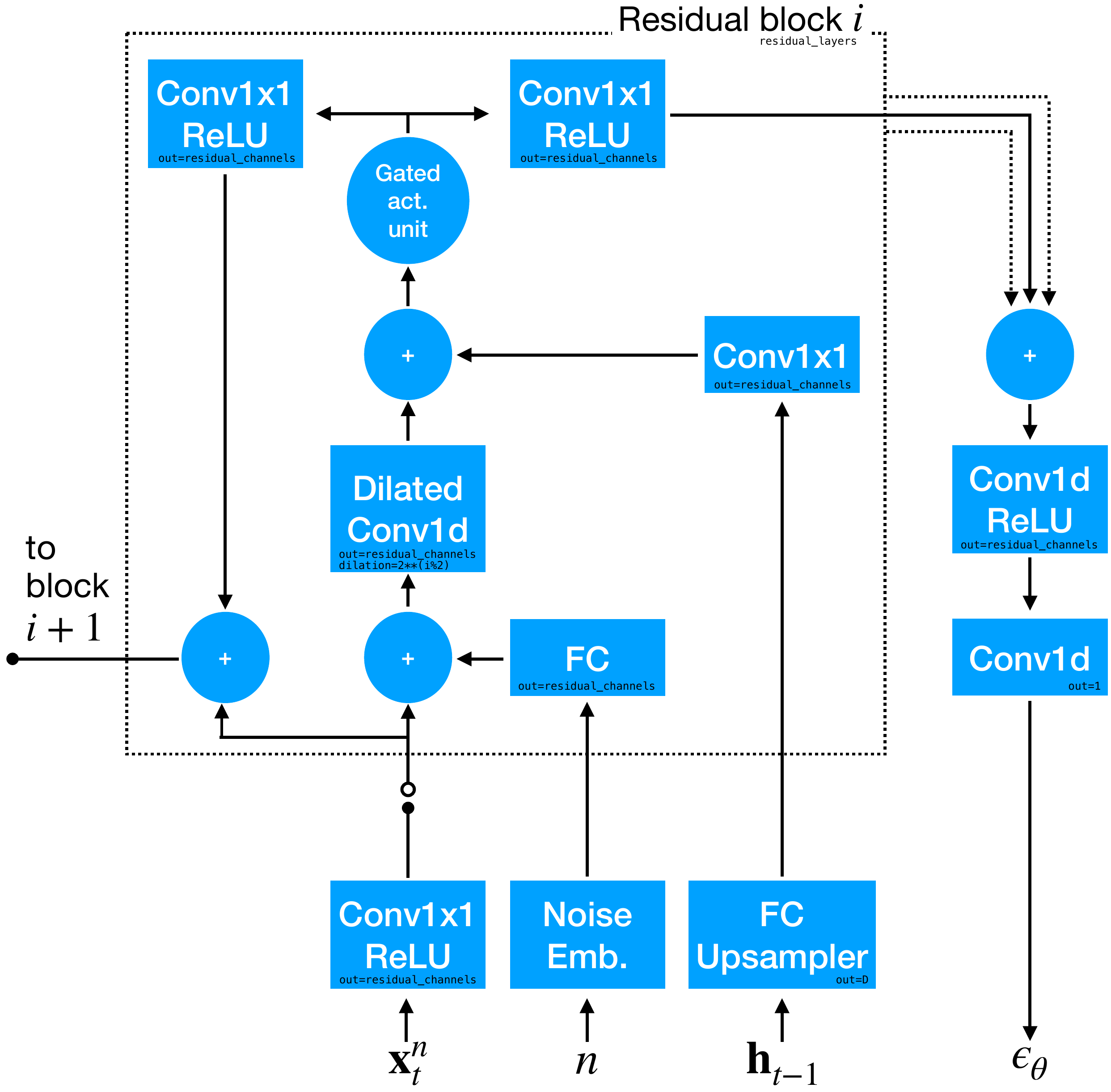}}
\caption{The network architecture of $\epsilon_\theta$ consisting of  $\mathtt{residual\_layers}=8$ conditional residual blocks with the Gated Activation Unit $\sigma(\cdot) \odot \tanh(\cdot)$ from \cite{NIPS2016_b1301141}; whose skip-connection outputs are summed up to compute the final output.  \texttt{Conv1x1} and \texttt{Conv1d} are 1D convolutional layers with filter size of $1$ and $3$, respectively,  circular padding so that the spatial size remains $D$, and all but the last convolutional layer has output channels $\mathtt{residual\_channels}=8$. \texttt{FC} are linear layers used to up/down-sample the input to the appropriate size for broadcasting.} 
\label{epsilon-theta}
\end{center}
\vskip -0.2in
\end{figure}

The RNN consists of $2$ layers of an LSTM with the hidden state $\mathbf{h}_t \in \mathbb{R}^{40}$ and we encode the noise index $n \in \{1, \ldots, N \}$ using the Transformer's \cite{NIPS2017_7181} Fourier positional embeddings, with $N_{\max} = 500$,  into  $ \mathbb{R}^{32}$ vectors.   The network $\epsilon_\theta$ consists of conditional 1-dim dilated ConvNets with residual connections adapted from the \texttt{WaveNet} \cite{45774} and \texttt{DiffWave} \cite{diffwave} models. Figure \ref{epsilon-theta} shows the schematics of a single residual block $i=\{0, \ldots, 7\}$ together with the final output from the sum of all the $8$ skip-connections. All, but the last, convolutional network layers have an output channel size of $8$ and we use a \emph{bidirectional} dilated convolution in each block $i$ by setting its dilation to $2^{i \% 2}$. We use a validation set from the training data of the same size as the test set to tune the number of epochs for early stopping.

All experiments run on a single Nvidia V100 GPU with $16$GB of memory.

\subsection{Results}

\begin{table*}[t]
\caption{Test set $\mathrm{CRPS}_{\mathrm{sum}}$ comparison (lower is better) of models on six real world data sets. Mean and standard error metrics for \texttt{TimeGrad} obtained by re-training and evaluating $10$ times.}
\label{crps-sum}
\vskip 0.15in
\begin{center}
\begin{tabular}{ccccccc}
\toprule
Method & \texttt{Exchange} & \texttt{Solar} &  \texttt{Electricity} & \texttt{Traffic}  & \texttt{Taxi}    & \texttt{Wikipedia}\\
\midrule 
\texttt{VES} & $ \mathbf{0.005} \scriptstyle{\pm 0.000}$ & $0.9 \scriptstyle{\pm 0.003}$ & $0.88 \scriptstyle{\pm 0.0035}$ & $0.35 \scriptstyle{\pm 0.0023}$ & - & -\\
 \hline

\texttt{VAR} & $ \mathbf{0.005}  \scriptstyle{\pm 0.000}$ & $0.83 \scriptstyle{\pm 0.006}$ & $0.039 \scriptstyle{\pm 0.0005}$ & $0.29 \scriptstyle{\pm 0.005}$ &  - & -   \\
 \hline

\texttt{VAR-Lasso} &  $0.012 \scriptstyle{\pm 0.0002}$ & $0.51 \scriptstyle{\pm 0.006}$ & $0.025 \scriptstyle{\pm 0.0002}$ &  $0.15 \scriptstyle{\pm 0.002}$ & - & $3.1 \scriptstyle{\pm 0.004}$   \\
 \hline

\texttt{GARCH} & $0.023 \scriptstyle{\pm 0.000}$ & $0.88 \scriptstyle{\pm 0.002}$ & $0.19 \scriptstyle{\pm 0.001}$ &    $0.37 \scriptstyle{\pm 0.0016}$ & - & - \\
 \hline

\texttt{KVAE} & $0.014 \scriptstyle{\pm 0.002}$ &  $0.34 \scriptstyle{\pm 0.025}$ & $0.051 \scriptstyle{\pm 0.019}$ &   $0.1 \scriptstyle{\pm 0.005}$ & - & $0.095 \scriptstyle{\pm 0.012}$   \\
 \hline

\makecell{\texttt{Vec-LSTM} \\ \texttt{ind-scaling}} & $0.008 \scriptstyle{\pm 0.001}$ &  $0.391 \scriptstyle{\pm 0.017}$ & $0.025 \scriptstyle{\pm 0.001}$ & $0.087 \scriptstyle{\pm 0.041}$ & $0.506 \scriptstyle{\pm 0.005}$ &  $0.133 \scriptstyle{\pm 0.002}$      \\
 \hline

\makecell{\texttt{Vec-LSTM} \\ \texttt{lowrank-Copula}} & $0.007 \scriptstyle{\pm 0.000}$ &  $0.319 \scriptstyle{\pm 0.011}$ & $0.064 \scriptstyle{\pm 0.008}$ & $0.103 \scriptstyle{\pm 0.006}$ &  $0.326 \scriptstyle{\pm 0.007}$ & $0.241 \scriptstyle{\pm 0.033}$    \\
 \hline

\makecell{ \texttt{GP} \\ \texttt{scaling}} & $0.009 \scriptstyle{\pm 0.000}$ & $0.368 \scriptstyle{\pm 0.012}$ & $0.022 \scriptstyle{\pm 0.000}$ & $0.079 \scriptstyle{\pm 0.000}$ &  $0.183 \scriptstyle{\pm 0.395}$ &    $1.483 \scriptstyle{\pm 1.034}$  \\
 \hline

\makecell{\texttt{GP} \\ \texttt{Copula}} & $0.007 \scriptstyle{\pm 0.000}$ &  $0.337 \scriptstyle{\pm 0.024}$ & $0.0245 \scriptstyle{\pm 0.002}$ &  $0.078 \scriptstyle{\pm 0.002}$ &  $0.208 \scriptstyle{\pm 0.183}$ &  $0.086 \scriptstyle{\pm 0.004}$  \\
 \hline

 \makecell{\texttt{Transformer}\\ \texttt{MAF}} & $ \mathbf{0.005}  \scriptstyle{\pm 0.003}$ & $0.301 \scriptstyle{\pm 0.014}$ & $0.0207 \scriptstyle{\pm 0.000}$ & $0.056 \scriptstyle{\pm 0.001}$ & $0.179 \scriptstyle{\pm 0.002}$ & $0.063 \scriptstyle{\pm 0.003}$        \\
  \hline

 \textbf{\texttt{TimeGrad}} & $ 0.006  \scriptstyle{\pm 0.001}$ & 
 $ \mathbf{0.287} \scriptstyle{\pm 0.02}$  & $\mathbf{0.0206} \scriptstyle{\pm 0.001}$ & $\mathbf{0.044} \scriptstyle{\pm 0.006}$ & $\mathbf{0.114} \scriptstyle{\pm 0.02} $ & $\mathbf{0.0485} \scriptstyle{\pm 0.002} $ \\
\bottomrule
\end{tabular}
\end{center}
\vskip -0.1in
\end{table*}

Using the $\mathrm{CRPS}_{\mathrm{sum}}$ as an evaluation metric, we compare test time predictions of \texttt{TimeGrad} to a wide range of existing methods including  classical multivariate methods:
\begin{itemize}
\item \texttt{VAR} \citep{luetkepohl2007new} a mutlivariate linear vector auto-regressive model with lags corresponding to the periodicity of the data,
\item \texttt{VAR-Lasso} a Lasso regularized \texttt{VAR},
\item \texttt{GARCH} \citep{RePEc:jae:japmet:v:17:y:2002:i:5:p:549-564} a multivariate conditional heteroskedastic model and 
\item \texttt{VES} a innovation state space model \citep{HyndmanKoehler};
\end{itemize}
as well as deep learning based methods namely:
\begin{itemize}
\item \texttt{KVAE} \citep{NIPS2017_7b7a53e2} a variational autoencoder to represent the data on top of a linear state space model which describes the dynamics,
\item \texttt{Vec-LSTM-ind-scaling}  \cite{NIPS2019_8907} which models the dynamics via an RNN and outputs the parameters of an \emph{independent} Gaussian distribution with mean-scaling,
\item \texttt{Vec-LSTM-lowrank-Copula} \cite{NIPS2019_8907} which instead parametrizes a low-rank plus diagonal covariance via Copula process,
\item \texttt{GP-scaling} \cite{NIPS2019_8907} which unrolls an LSTM with scaling  on each individual time series before reconstructing the joint distribution via a low-rank Gaussian,  
\item \texttt{GP-Copula} \cite{NIPS2019_8907} which unrolls an LSTM on each individual time series and then the joint emission distribution is given by  a low-rank plus diagonal covariance Gaussian copula and
\item \texttt{Transformer-MAF} \cite{temp-flow} which uses Transformer  \cite{NIPS2017_7181}  to model the temporal conditioning and Masked Autoregressive Flow \cite{Papamakarios:2017:maf} for the distribution emission model. 
\end{itemize}

Table~\ref{crps-sum} lists the corresponding  $\mathrm{CRPS}_{\mathrm{sum}}$ values averaged over $10$ independent runs together with their empirical standard deviations and shows that the \texttt{TimeGrad} model sets the new state-of-the-art on all but the smallest of the benchmark data sets. Note that  flow based models must apply continuous transformations onto a continuously connected distribution, making it difficult to model disconnected modes. Flow models assign spurious density to connections between these modes leading to potential inaccuracies. Similarly the generator network in variational autoencoders must learn to map from some continuous space to a possibly disconnected space which might not be possible to learn. In contrast  EMBs do not suffer from these issues \cite{NEURIPS2019_378a063b}.

\subsection{Ablation}

\begin{figure}[ht]
\vskip 0.2in
\begin{center}
\centerline{\includegraphics[width=\columnwidth]{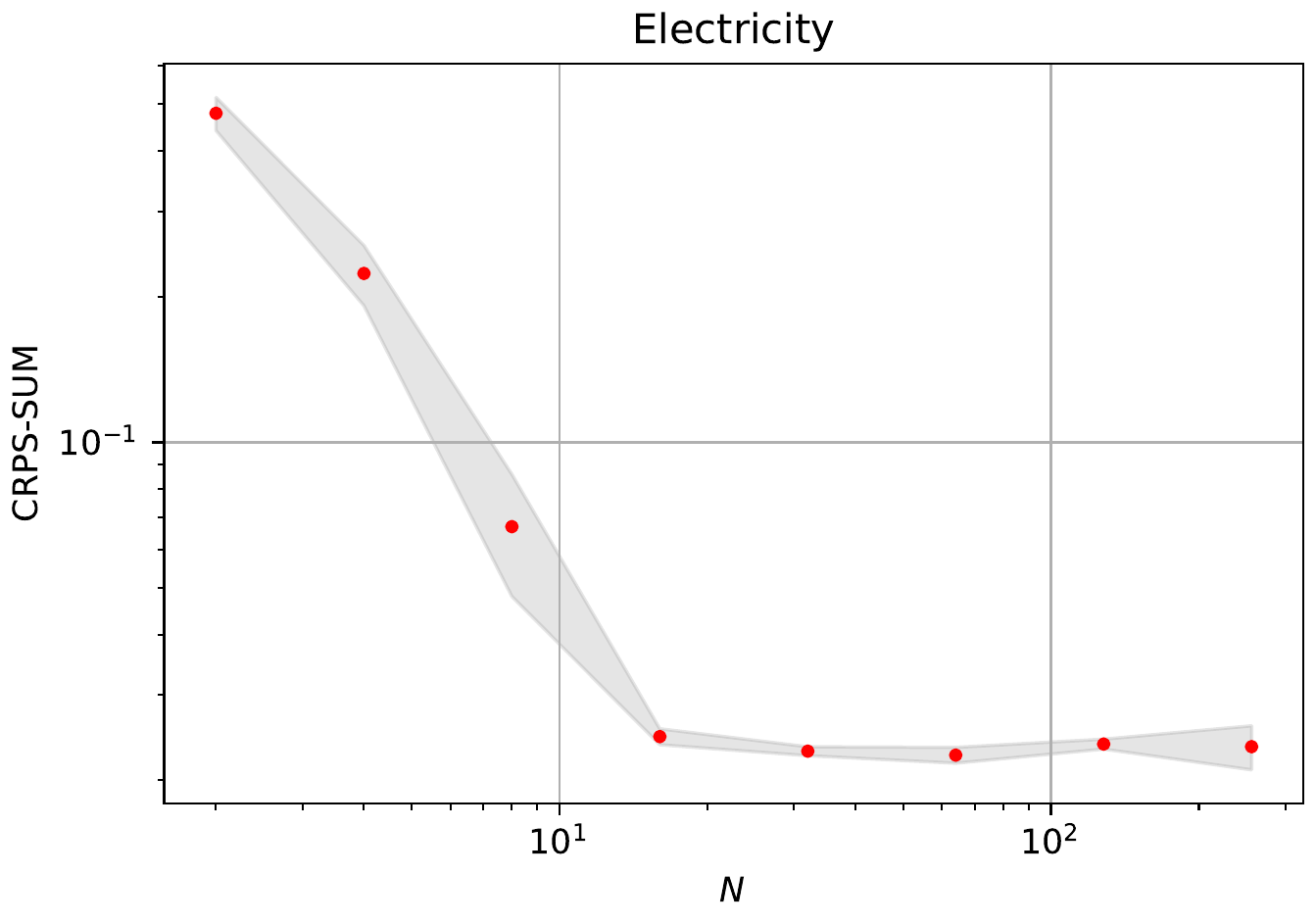}}
\caption{\texttt{TimeGrad} test set $\mathrm{CRPS}_{\mathrm{sum}}$ for \texttt{Electricity} data by varying total diffusion length $N$. Good performance is established already at $N \approx 10$ with optimal value at $N\approx100$. The mean and standard errors obtained over $5$ independent runs. We see similar behaviour with other data sets.
}
\label{ablation}
\end{center}
\vskip -0.2in
\end{figure}

The length $N$ of the forward process is a crucial hyperparameter, as a bigger $N$ allows the reverse process to be approximately Gaussian \cite{pmlr-v37-sohl-dickstein15} which assists the Gaussian parametrization (\ref{eqn:parametrization}) to approximate it better.  We evaluate  to which extent, if any at all, larger $N$ affects  prediction performance, with  an ablation study where we record the test set $\mathrm{CRPS}_{\mathrm{sum}}$ of the \texttt{Electricity} data set for different total diffusion process lengths $N = 2, 4, 8, \ldots, 256$  while keeping all other hyperparemeters unchanged. The results are then plotted  in Figure~\ref{ablation} where  we note that $N$ can be reduced down to $\approx10$ without significant performance loss. An optimal value is achieved at $N\approx100$ and larger levels are not beneficial if all else is kept fixed.

\begin{figure}[th!]
\vskip 0.2in
\begin{center}
\centerline{\includegraphics[width=\columnwidth]{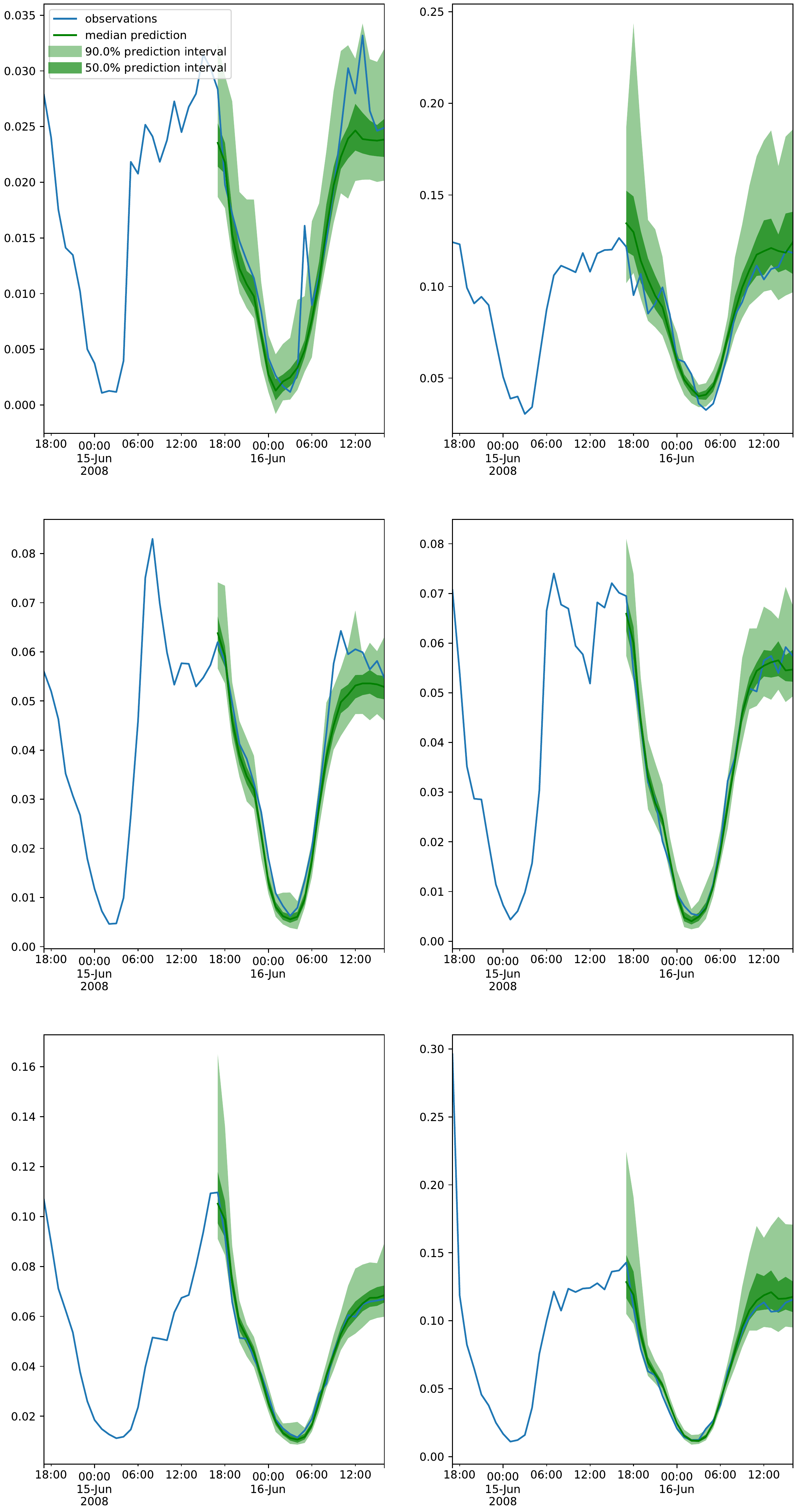}}
\caption{\texttt{TimeGrad} prediction intervals and test set ground-truth for \texttt{Traffic} data of the first $6$ of $963$ dimensions from first rolling-window. Note that neighboring entities have an order of magnitude difference in scales.}
\label{traffic-prediction}
\end{center}
\vskip -0.2in
\end{figure}

To highlight the predictions of \texttt{TimeGrad} we  show in Figure~\ref{traffic-prediction} the predicted median, $50\%$ and $90\%$ distribution intervals of the first $6$ dimensions of the full $963$ dimensional multivariate forecast of the \texttt{Traffic} benchmark.

\section{Related Work}
\label{sec:Related}

\subsection{Energy-Based Methods}

The EBM of \cite{ho2020denoising} that we adapt is based on methods that learn the gradient of the log-density with respect to the \emph{inputs}, called Stein Score function \cite{JMLR:v6:hyvarinen05a, doi:10.1162}, and at inference time use this gradient estimate via Langevin dynamics to sample from the model of this complicated data distribution  \cite{NEURIPS2019_3001ef25}. These models achieve impressive results for image generation \cite{ho2020denoising, NEURIPS2020_song_erman} when trained in an unsupervised fashion without requiring adversarial optimization. By perturbing the data using multiple noise scales, the learnt Score network captures both coarse and fine-grained data features. 

The closest related work to \texttt{TimeGrad} is in the recent non-autoregressive conditional methods for high fidelity waveform generation \cite{wavegrad, diffwave}. Although these methods learn the distribution of vector valued data via denoising diffusion methods, as done here, they do not consider its temporal development. Also neighboring dimensions of waveform data are highly correlated and have a uniform scale,  which is not necessarily true for multivariate time series problems where neighboring entities occur arbitrarily (but in a fixed order) and can have different scales.
\cite{NEURIPS2019_378a063b} also use EBMs to model one and multiple steps for a trajectory modeling task in an non-autoregressive fashion. 

\subsection{Time Series Forecasting}

Neural time series methods have recently become  popular ways of solving the prediction problem via univariate point forecasting methods  \cite{oreshkin2020nbeats, SMYL202075} or univariate probabilistic methods \cite{DBLP:journals/corr/FlunkertSG17}. In the multivariate setting we also have point forecasting methods \cite{Lai:2018:MLS:3209978.3210006, NIPS2019_8766} as well as probabilistic methods, like this method,  which explicitly model the data distribution using Gaussian copulas \cite{NIPS2019_8907}, GANs \cite{NEURIPS2019_c9efe5f2}, or normalizing flows \cite{NIPS2020_Emmanuel, temp-flow}. Bayesian neural networks can also be used to provide \emph{epistemic} uncertainty in forecasts as well as detect distributional shifts \cite{8215650}, although these methods often do not perform as well empirically \cite{pmlr-v119-wenzel20a}.

\section{Conclusion and Future Work}
\label{sec:Conclusion}

We have presented \texttt{TimeGrad}, a versatile multivariate probabilistic time series forecasting method that leverages the exceptional performance of EBMs to learn and sample from the  distribution  of the next time step,  autoregressivly. Analysis of \texttt{TimeGrad} on six commonly used time series  benchmarks establishes the new state-of-the-art  against competitive methods.

We note that while training \texttt{TimeGrad} we do not need to loop over the EBM function approximator $\epsilon_\theta$, unlike in the normalizing flow setting where we have multiple stacks of bijections. However while sampling we do loop $N$ times over $\epsilon_\theta$. A possible strategy to improve sampling times introduced in \cite{wavegrad} uses a combination of improved variance schedule and an $L_1$ loss to allow sampling with fewer steps at the cost of a small reduction in quality if such a trade-off is required. A recent paper \cite{implicit-denoise} generalize the diffusion processes via a class of non-Markovian processes which also allows for faster sampling.

The use of normalizing flows for discrete valued data dictates that one dequantizes it \cite{Theis2015d}, by adding uniform noise to the data, before using the flows to learn. Dequantization is not needed  in the EBM setting and future work could explore methods of explicitly modeling discrete distributions. 

As noted in \cite{NEURIPS2019_378a063b} EBMs exhibit better out-of-distribution (OOD) detection than other likelihood models. Such a task requires models to have a high likelihood on the data manifold and low at all other locations. Surprisingly \cite{nalisnick2018do} showed that likelihood models, including flows, were assigning higher likelihoods to OOD data whereas EBMs do not suffer from this issue since they penalize  high probability under the model but low probability under the data distribution  explicitly. Future work could evaluate the usage of \texttt{TimeGrad} for anomaly detection tasks. 

For long time sequences, one could replace the RNN with a Transformer architecture  \cite{temp-flow} to provide better conditioning for the EBM emission head. Concurrently, since EBMs are not constrained by the form of their functional approximators, one natural way to improve the model would be to incorporate architectural choices that best encode the inductive bias of the problem being tackled, for example with graph neural networks \cite{DBLP:conf/aistats/NiuSSZGE20} when the relationships between entities are known.

\bibliography{references}
\bibliographystyle{icml2021}

\end{document}